\documentclass[a4paper,11pt]{article}

\usepackage{booktabs}

\usepackage{lineno}
\usepackage[colorlinks=false]{hyperref}
\usepackage{bm}
\usepackage{graphicx}
\usepackage{amssymb}
\usepackage{amsmath}
\usepackage{tabularx}
\usepackage{natbib}
\usepackage{floatrow}
\usepackage{caption}
\usepackage{multirow}
\usepackage{amsmath}
\usepackage{amsfonts}
\usepackage{amssymb}
\usepackage{xcolor}
\usepackage{subfig}
\usepackage{graphicx,amsmath,bm}
\usepackage{siunitx}
\usepackage{array}
\usepackage{authblk}
\usepackage{rotating}
\usepackage{enumerate}
\usepackage{ragged2e}

\usepackage[left=15mm,right=15mm,top=1.5cm,bottom=1.5cm,includeheadfoot]{geometry}
\setcitestyle{square,numbers}
\begin{document}
	
	\title{Improved Physics-Driven Neural Network to Solve Inverse Scattering Problems}
	
	\author[1]{Yutong~Du}
	\author[1]{Zicheng~Liu}
	\author[2]{Bo~Wu}
	\author[3]{Jingwei~Kou}
	\author[4]{Hang~Li}
	\author[1]{Changyou~Li}
	\author[1]{Yali~Zong}
    \author[1]{Bo~Qi}
	
	\affil[1]{\scriptsize Department of Electronic Engineering, School of Electronics and Information, Northwestern Polytechnical University, Xi'an 710029, China}
	\affil[2]{\scriptsize China North Communication Technology Co., Ltd., Xinxiang 453000, China}
	\affil[3]{\scriptsize The Advanced Optical Instrument Research Department, Xi’an Institute of Optics and Precision Mechanics, Chinese Academy of Sciences, Xi'an 710119, China}
    \affil[4]{\scriptsize Hangzhou Puhe Technology Co., Ltd., Hangzhou 311113, China}
	\maketitle
	
	\abstract{
		This paper presents an improved physics-driven neural network (IPDNN) framework for solving electromagnetic inverse scattering problems (ISPs). A new Gaussian-localized oscillation-suppressing window (GLOW) activation function is introduced to stabilize convergence and enable a lightweight yet accurate network architecture. A dynamic scatter subregion identification strategy is further developed to adaptively refine the computational domain, preventing missed detections and reducing computational cost. Moreover, transfer learning is incorporated to extend the solver’s applicability to practical scenarios, integrating the physical interpretability of iterative algorithms with the real-time inference capability of neural networks. Numerical simulations and experimental results demonstrate that the proposed solver achieves superior reconstruction accuracy, robustness, and efficiency compared with existing state-of-the-art methods.}
	
	\section{Introduction}
	%
	%
	%
	%
	Electromagnetic inverse scattering problems (ISPs) \cite{chen2018computationalEMIS} aim to reconstruct the shape, size, and electromagnetic properties of unknown objects. They play a crucial role in applications such as nondestructive testing of composite materials \cite{liu2017electromagnetic,li20213}, ground-penetrating radar \cite{Dai2023GPR3D,Zhang2025GPR}, and security screening \cite{Ahmed2021MicroSec,Murug2024Air}. Due to their inherent ill-posedness, ISPs are highly sensitive to measurement noise, which can lead to unstable imaging results. The nonlinear mapping between the scattered field and the scatterers often causes solutions to fall into local optima. Moreover, ISPs are typically underdetermined, as the limited measured data are insufficient to uniquely determine the spatial distribution of targets. To address these challenges, numerous computational strategies have been developed. 
	
	Classical solvers can be broadly categorized into non-iterative and iterative methods. Non-iterative methods, including Born approximation (BA) \cite{gao2006BA,habashyi1993BA}, backpropagation (BP) \cite{devaney1982BP,tsili1998BP}, Rytov approximation (RA) \cite{devaney1981RA,slaney1984RA,alon1993RA}. With linearization assumptions, these methods can achieve fast reconstruction but fail to account for multiple-scattering effects, making them unsuitable for reconstructing high-contrast scatterers. Iterative methods, including Born iterative method (BIM) \cite{wang1989BIM,sung1999BIM}, distorted Born iterative method (DBIM) \cite{chew1990DBIM,haddadin1995DBIM}, contrast source inversion (CSI) \cite{peter1997CSI,richard2001CSI} and subspace optimization (SOM) \cite{chen_2010SOM,chen2010SOM,pan2011SOM}. Considering multiple-scattering behaviors, these methods can handle a wider range of scatterers. However, they suffer from high computational cost and limited reconstruction efficiency, which hinder the real-time applications.

	Recent advances in deep learning have led to significant progress in solving ISPs. Existing deep-learning-based solvers can be divided into data-driven \cite{wei2018BPS,Li2018DeepNIS} and physics-constrained \cite{wei2019PhaNN,Liu2022PhaGuiNN,Liu2022SOMnet,Tao2023NNBIM,2023ZhangCSINet,Du2025QuaDNN} models. Data-driven approaches act as ``black box", learning only an input–output mapping without enforcing physical constraints, which limits their generalization ability. Physics-constrained models, in contrast, incorporate physical information into the loss function, improving both accuracy and generalization. Nevertheless, their dependence on large training datasets restricts broader applicability.
	
	To overcome these limitations, the physics-driven neural network (PDNN) framework \cite{Du2025PDNN} was introduced by the presenting authors. PDNNs optimize the network solely based on measured scattered-field data and electromagnetic physical laws. This approach avoids the generalization problem while leveraging the strong fitting capability of neural networks to achieve high-precision reconstructions.
	
	In this paper, we propose an improved physics-driven neural network (IPDNN), which enhances the original PDNN  \cite{Du2025PDNN} in three key aspects:
	
    \begin {enumerate}

    \item GLOW activation function: A Gaussian-localized oscillation suppressing window activation is designed to improve convergence stability and allow a lightweight network while maintaining high accuracy.
    
    \item Dynamic scatter subregion identification: A dynamic identification strategy is incorporated during training to rediscover overlooked subregions and improve computational efficiency by focusing only on relevant regions.
    
    \item Transfer learning: Transfer learning is introduced to extend the applicability of the IPDNN to practical tasks, combining the interpretability of physics-based iterative methods with the real-time inference of neural networks.
    
    \end {enumerate}
	
	The remainder of this paper is organized as follows. Section\ref{sec:formulateISPs} illustrates the involved forward and inverse problems. Section\ref{sec:IPDNNsolver} presents the proposed activation function, dynamic subregion identification algorithm, and transfer-learning framework. Section\ref{sec:NumeSimu} provides numerical and experimental results to demonstrate the superiority of the proposed solver. Conclusions are made in Section\ref{sec:conclusions}.
	
	\section{Formulation of ISPs}
	\label{sec:formulateISPs}
	\begin{figure}[!t]
		\centering
		\includegraphics[width = 0.38\linewidth]{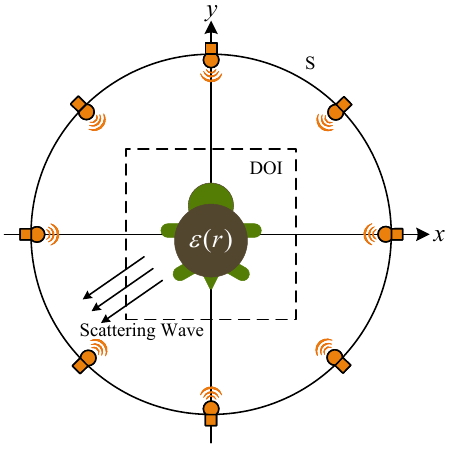}
		\caption{Sketch of the concerned two-dimensional inverse scattering problems.}
		\label{fig:ISPs}
	\end{figure}
	The configuration of the considered two-dimensional (2-D) problem is illustrated in Figure~\ref{fig:ISPs}. The background medium has relative permittivity $\epsilon_0$ and permeability $\mu_0$. Non-magnetic scatterers with translational symmetry along the $z$-axis are located within the domain of interest (DOI) and illuminated by a transverse-magnetic (TM) $z$-polarized wave \cite{meaney1995TM}. The scattered field is collected by receivers distributed along the observation surface $\textbf{S}$.
	
	The problem is governed by the state equation, which describes the coupling between the incident field and the scatterers, and the data equation, which relates the induced current to the measured scattered field. The state equation can be written as
    \begin{equation}
    	\mathbf{E}^\text{tot}(\mathbf{r}) = \mathbf{E}^\text{inc}(\mathbf{r}) + k_0^2\int_\text{DOI}g(\mathbf{r},\mathbf{r}^\prime)\mathbf{J}(\mathbf{r}^\prime)d\mathbf{r}^\prime,\mathbf{r}\in\text{DOI},
    	\label{eq:stateEqu}
    \end{equation}
    where $\mathbf{E}^\text{tot}$ and $\mathbf{E}^\text{inc}$ denote the total and incident electric fields, respectively. $k_0$ is the background wavenumber. $g$ stand for the 2-D Green's function \cite{liu2025computational}. The induced current is given by $\mathbf{J}(\mathbf{r}^\prime) = (\epsilon_r(\mathbf{r}^\prime)-1)\mathbf{E}^\text{tot}(\mathbf{r}^\prime)$, where $\epsilon_r$ is the relative permittivity.
    
    The data equation expresses the measured scattered fields
    \begin{equation}
    	\mathbf{E}^\text{sca}_\text{mea}(\mathbf{r}) = k_0^2\int_\text{DOI}g(\mathbf{r},\mathbf{r}^\prime)\mathbf{J}(\mathbf{r}^\prime)d\mathbf{r}^\prime, \mathbf{r}\in\text{S}.
    	\label{eq:dataEqu}
    \end{equation} 
    
    Inverse scattering problems aims at retrieving the spatial distribution of $\epsilon_r$ from the measured scattered field $\mathbf{E}^\text{sca}_\text{mea}$. 
    
    Due to the inherent ill-posedness of ISPs, regularization techniques\cite{Oliveri2017Regular, Leonid1992TV} can be applied to incorporate prior knowledge and impose constraints on the solution to improve the reconstruction stability. The optimization problem for estimating $\hat{\epsilon_r}$ is formulated as
    \begin{equation}
    	L(\hat{\boldsymbol{\epsilon}}_r) =||{\mathbf{E}}^\text{sca}_\text{mea}-\hat{\mathbf{E}}^\text{sca}_\text{mea}(\hat{\boldsymbol{\epsilon}}_r)||_2 +  g(\hat{\boldsymbol{\epsilon}}_r),
    	\label{eq:optimizationEq}
    \end{equation} 
    where $\hat{\mathbf{E}}^\text{sca}_\text{mea}(\hat{\boldsymbol{\epsilon}}_r)$ is the synthetic field corresponding to the solution $\hat{\boldsymbol{\epsilon}}_r$ and computed with method of moments (MoM) \cite{Ney1985MoM,LAKHTAKIA1992MoM,Gibson2021MoM} in the paper. $g(\hat{\boldsymbol{\epsilon}}_r)$ is the regularization term.
	
	\section{Improved Physics-Driven Neural Network}
    \label{sec:IPDNNsolver}
    This section presents the proposed improved physics-driven neural network (IPDNN) solver. The IPDNN employs a fully connected network architecture combined with a newly designed Gaussian-localized oscillation-suppressing window (GLOW) activation function. The GLOW function enhances convergence stability and accuracy while reducing network complexity. Furthermore, a dynamic scatter subregion identification mechanism is incorporated to mitigate missed detections during inversion, and transfer learning is introduced to extend the method's applicability to real-world scenarios.

	\subsection{Neural network settings}
    \subsubsection{GLOW activation function}
    \label{subsec:ActFun}
    
    \begin{figure}[!t]
    	\centering
    	\includegraphics[width = .5\linewidth]{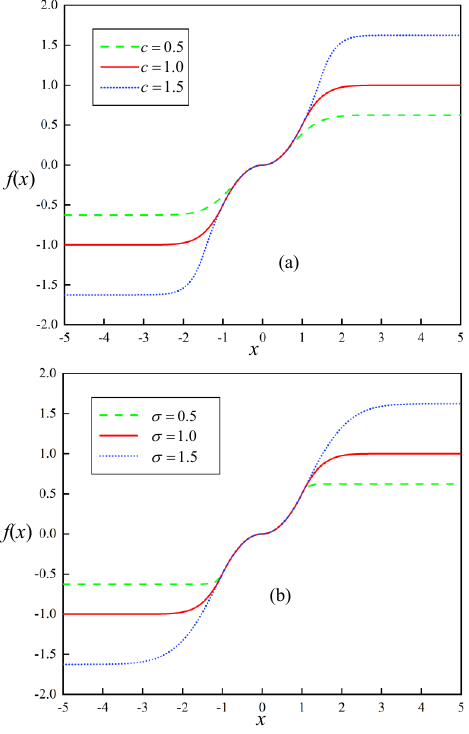}
    	\caption{Variation of GLOW function value when (a) fixing $\sigma=1$ and changing $c$ and (b) fixing $c=1$ and changing $\sigma$.}
    	\label{fig:CandSigma}
    \end{figure}

    Inverse scattering problems involve complex nonlinear mappings between scattered fields and spatial permittivity distributions. To address this, we propose a Gaussian-localized oscillation-suppressing window (GLOW) activation function defined as

    \begin{equation}
    	\varphi(x) = \text{sign}(x)\times\left\{
    	\begin{array}{ll}
    		\frac{x^2}{2}, & \left| x \right| \leq c, \\
    		\frac{\sigma^2}{2}(1-e^{\frac{c^2-x^2}{\sigma^2}}), & \left| x \right| > c,
    	\end{array}
    	\right.
    \end{equation} 
    where $c$ and $\sigma$ are trainable and their effects can be observed from Figure~\ref{fig:CandSigma}. The proposed GLOW function possesses the following desirable properties:
    \begin {enumerate}
    \item[--] Nonlinearity: Enables the network to approximate complex mappings.
    
    \item[--] Differentiability: Ensures smooth gradient propagation during training.
    
    \item[--] Unsaturation: Prevents gradient vanishing in deep networks.
    
    \item[--] Monotonicity: Facilitates convergence to the global optimum.
    
    \item[--] Boundedness: Improves stability when processing large inputs.
    
    \end {enumerate}

    The derivative of the GLOW function is expressed as 
    \begin{equation}
    \varphi^\prime(x) = \left\{
    \begin{array}{cl}
         \left| x \right|, & \left| x \right| \leq c \\
         \left| x \right|e^{\frac{c^2-x^2}{\sigma^2}}, & \left| x \right| > c
    \end{array}
    \right.
    \end{equation}
    which reveals that $\varphi^\prime(x)$ increases linearly for small inputs but decays exponentially for large ones.This characteristic enables the function to preserve subtle features while suppressing the effect of abnormal samples with large amplitudes, leading to improved reconstruction stability and precision.
    
    \subsubsection{Neural network architecture}
    
    \begin{figure}[!t]
    	\centering
    	\includegraphics[width = .7\linewidth]{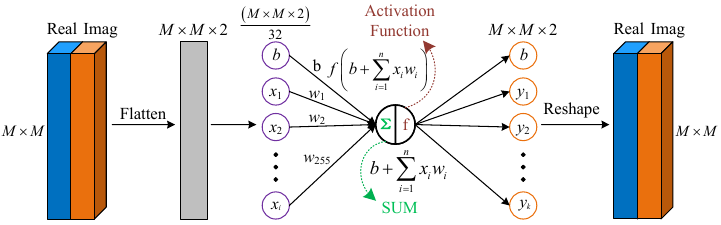}
    	\caption{The schematic diagram of the applied neural network architecture.}
    	\label{fig:FCN}
    \end{figure}
    As sketched in Figure~\ref{fig:FCN}, the IPDNN adopts a fully connected network architecture incorporating the proposed GLOW activation. Considering that scatterers may be lossy, both the input and output contain two channels representing the real and imaginary parts of the relative permittivity. This compact design leverages the expressive capability of the activation function to reduce network complexity without sacrificing accuracy.
    
    \subsubsection{Loss function}
    \label{subSec:lossFun}
    
    The total loss function used for training is a summation of three terms:
    \begin{equation}
    	\text{Loss} = L^{\text{Data}}+\alpha L^{\text{Bound}}+\beta L^{\text{TV}}
    	\label{LossFunc}
    \end{equation}
    where the data consistency term 
    \begin{equation}
    	L^{\text{Data}}=||{\mathbf{E}}^\text{sca}_\text{mea}-\hat{\mathbf{E}}^\text{sca}({\hat{\boldsymbol{\epsilon}}_r})||_2,
    	\label{eq:LData}
    \end{equation}
    is to ensures consistency between the measured and predicted scattered fields, the boundary constraint term
    \begin{equation}
    	L^{\text{Bound}}=\text{ReLU}(1-\mathbf{Re\{\hat{\boldsymbol{\epsilon}}_r\}}),
    	\label{eq:LreEp}
    \end{equation}
    constrains the reconstructed permittivity to physically meaningful values, and the total variation (TV) regularization term
    \begin{equation}
    	L^{\text{TV}}=\sum_{i,j}\sqrt{({\epsilon}_{i,j-1}-{\epsilon}_{i,j})^2+({\epsilon}_{i+1,j}-{\epsilon}_{i,j})^2}.
    	\label{eq:LTV}
    \end{equation}
    enforces spatial smoothness \cite{ma2023inverse}, where $\alpha$ and $\beta$ are weighting coefficients balancing the three terms.
    
    \begin{figure}[!ht]
    	\centering
    	\includegraphics[width = 0.38\linewidth]{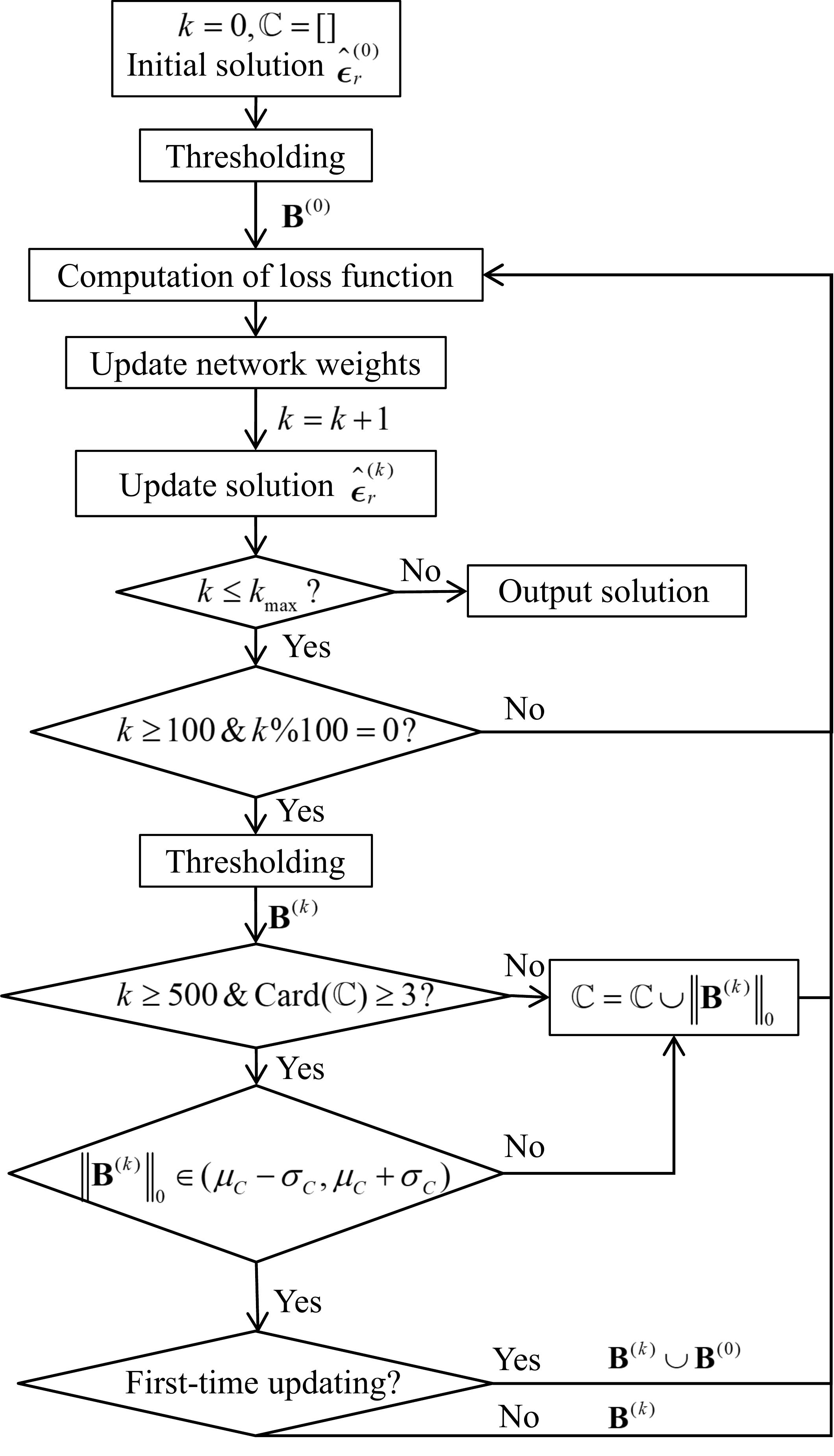}
    	\caption{The flow chart for the dynamic scatter subregion identification method.}
    	\label{fig:Flowchart}
    \end{figure}
    
    \subsection{Dynamic scatter subregion identification}
    \label{subSec:selectDOI}
    
    
    When evaluating $L^{\text{Data}}$, the scattered fields corresponding to the predicted permittivity $\hat{\boldsymbol{\epsilon}}_r$ need to be computed, and the computational complexity is proportional to the number of discretized grids within the domain of interest (DOI). In the original PDNN framework \cite{Du2025PDNN}, the scatter region was determined only once at initialization, which could lead to overlooked regions. To address this limitation, we introduce a dynamic scatter subregion identification mechanism that adaptively refines the region of interest during training.
    
    The process, depicted in Figure~\ref{fig:Flowchart}, begins with an initial estimate $\hat{\boldsymbol{\epsilon}}_r^{(0)}$ obtained using a data-driven U-Net solver. A binary map $\mathbf{B}^{(0)}$ is generated through thresholding:
    \begin{equation}
    	B(i,j) = \left\{
    	\begin{array}{cl}
    		1, & \text{if} \,\, \epsilon_r(i,j) \ge \mu+3\sigma \\
    		0, & \text{otherwise}
    	\end{array}
    	\right.
    \end{equation}
    where $\mu$ and $\sigma$ are the mean and the standard deviation of the smallest 30\% of the relative-permittivity values of the solution that input to the thresholding operator. 
    
    During iterative training, the neural network parameters are updated by minimizing the loss function while computing
    $\hat{\mathbf{E}}^\text{sca}$ only on the active grids (those with $B(i,j)=1$). The thresholding operation is repeated every 100 iterations, and updates to the subregion occur only after iteration $k\ge 500$ when the number of nonzero elements $C_\mathbf{B}$ becomes statistically stable. Stability is assessed by maintaining a set $\mathbb{C}$ of recent $C_\mathbf{B}$ values and verifying whether the current $C_\mathbf{B}$ lies within the range $(\mu_C-\sigma_C,\mu_C+\sigma_C)$, where $\mu_C$, $\sigma_C$ are the mean and the standard deviation of set $\mathbb{C}$. If this condition is met, the subregion is updated. When fist-time updating, the scatter region is updated using the union $\mathbf{B}^{(0)} \cup \mathbf{B}^{(k)}$ to ensure that the region of true scatterers is fully covered; Otherwise, updated as $\mathbf{B}^{(k)}$.
    
    This adaptive mechanism effectively refines the region of interest, preventing missed detections and improving computational efficiency.
    
    \subsection{Transfer learning}
    \label{subSec:TransferLearning}
    
    In many practical scenarios, such as nondestructive testing, the structural configuration and materials of an intact object are known. To leverage this prior information, transfer learning is incorporated into the IPDNN framework. Specifically, the network is first pretrained using data from the sound (defect-free) object. The learned weights are then transferred as initialization for the downstream defect-detection task. This initialization accelerates convergence, enhances imaging quality, and enables efficient adaptation to the new domain.
    
    The performance gain of this transfer learning strategy is demonstrated in Section\ref{subsec:ApTransLea}, where reductions in iteration count and the required number of transmitters are achieved.
    
    \subsection{Training settings and evaluation indicator}
    \label{subsec:trainSet}
    All ISP solvers in this study were implemented and trained on a workstation equipped with 128 GB RAM, a 3.2 GHz Intel i9 CPU and an NVIDIA GeForce RTX 4090 GPU. The network parameters are optimized using the Adam algorithm with an initial learning rate of 1$\times$10$^{-3}$. 
    
    The imaging performance was quantitatively assessed using the relative prediction error, defined as
    \begin{equation}
    	||{\boldsymbol{\epsilon}}_r-\hat{\boldsymbol{\epsilon}}_r||_\text{F}/||{\boldsymbol{\epsilon}}_r||_\text{F},
    \end{equation} 
    where $||\cdot||_\text{F}$ denotes the Frobenius norm.
    
    \section{Results and analysis}
    \label{sec:NumeSimu}
    
    \begin{figure*}
    	\centering
    	\includegraphics[width = \linewidth]{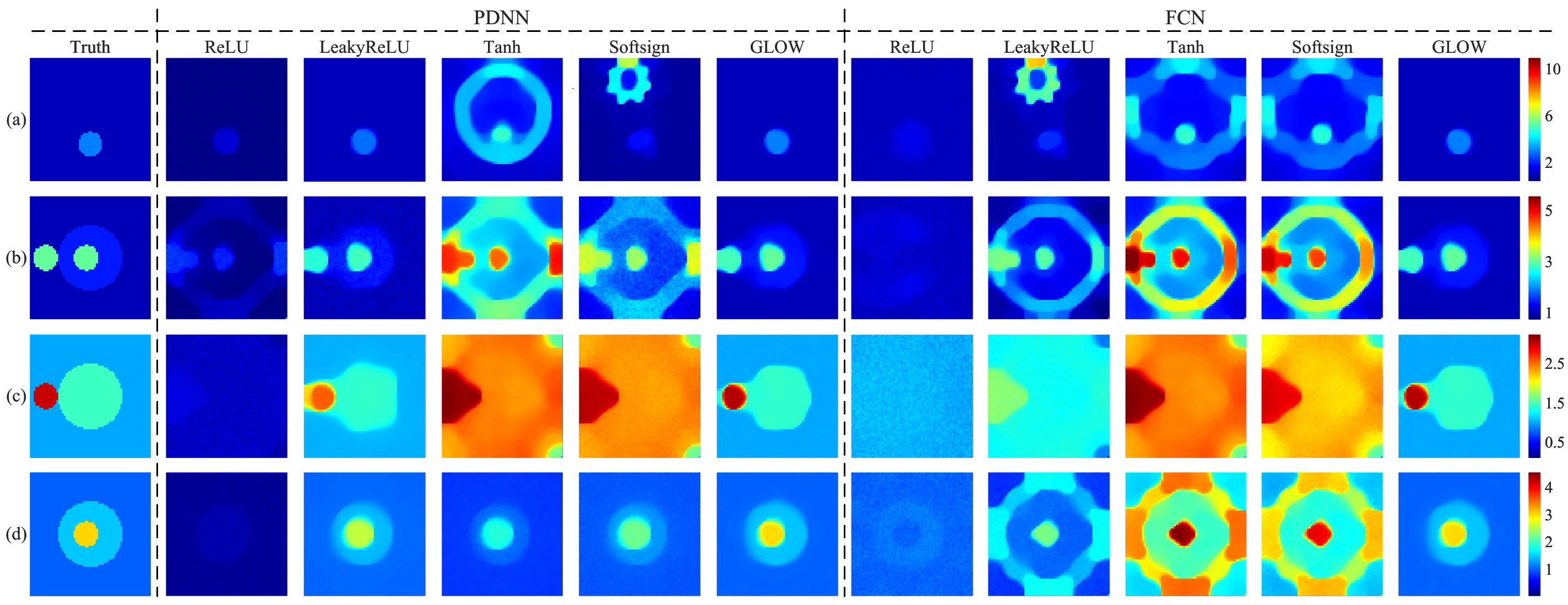}
    	\caption{Based on the experimental dataset (a) “dielTM\underline{~}dec8f”, (b) “FoamTwinDiel”, (c) “FoamDielExt” and (d) “FoamDielInt”, the imaging results from the solvers with PDNN \cite{Du2025PDNN} and fully connected layer (denoted by ``FCN") neural network architecture equipped with ReLU, LeakyReLU, Tanh, Softsign and GLOW activation function, respectively.}
    	\label{fig:diffActFunExp}
    \end{figure*}
    \begin{figure}[!t]
    	\centering
    	\includegraphics[width = 0.38\linewidth]{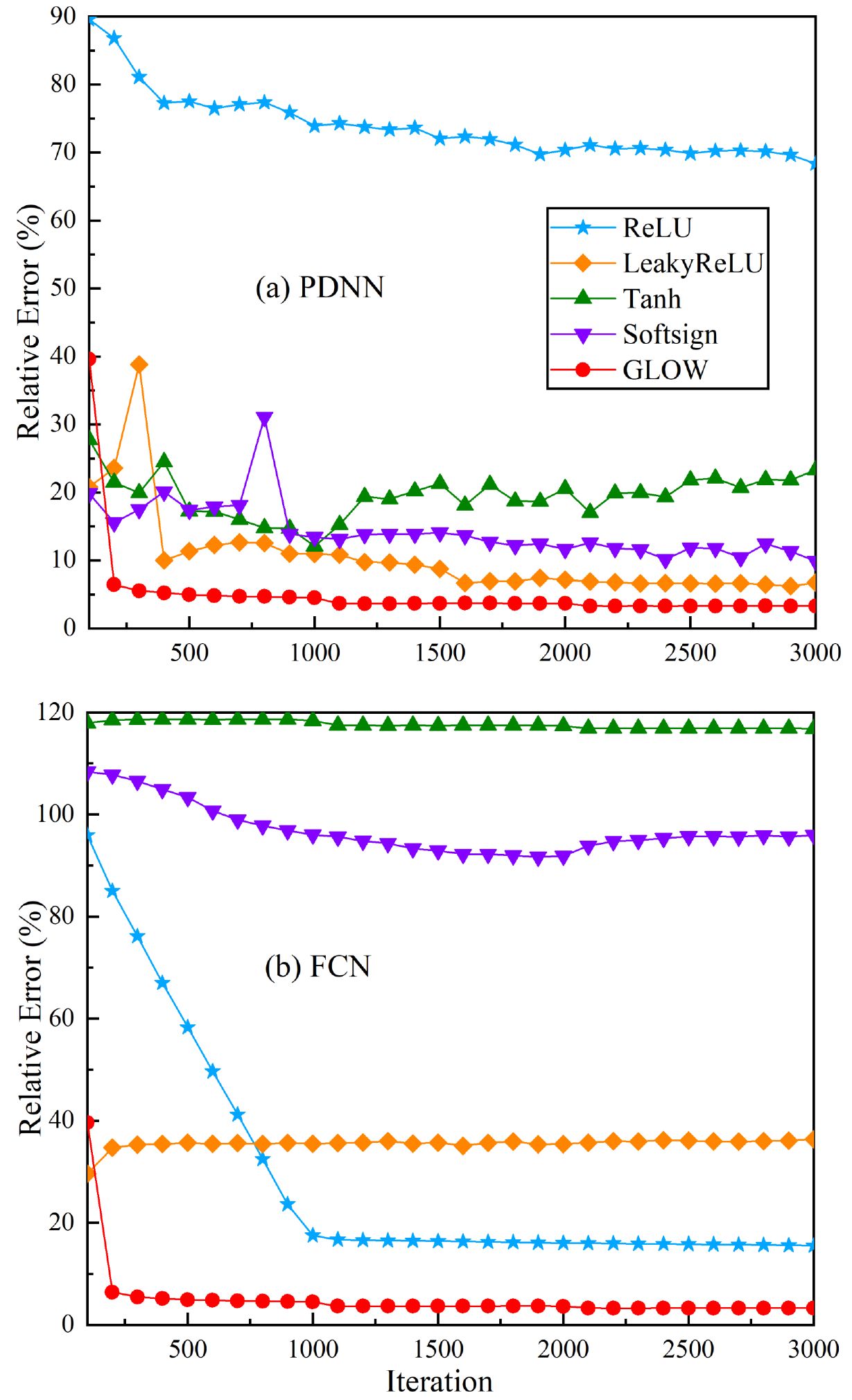}
    	\caption{The convergence curve corresponding to the imaging example of ``FoamDieInt" in Figure~\ref{fig:diffActFunExp}.}
    	\label{fig:diffActFun}
    \end{figure}
    Numerical and experimental studies are carried out to evaluate the proposed IPDNN framework. All numerical experiments are based on simulated data within a domain of interest (DOI) of 0.15m$\times$0.15m.  $36$ transmitters and $36$ receivers are uniformly distributed on the circle, which shares the same center with DOI and is with radius $20\lambda$, where $\lambda$ is the wavelength corresponding to the wave frequency 4 GHz. The DOI is discretized into $64\times 64$ grids. Experimental validations are performed using the Fresnel Institute datasets, with detailed measurement configurations available in \cite{Kamal2001Exp,geffrin2005Fresnel}.
    
    \subsection{Effects of GLOW activation function}
    \label{subsec:GLOW}

    To assess the effectiveness of the proposed GLOW activation, we compare it against conventional activation functions, including ReLU, LeakyReLU, Tanh, and Softsign. Two learning schemes are considered: the original PDNN framework \cite{Du2025PDNN} and the present IPDNN framework with a fully connected network (FCN) structure.
    
    Based on the experimental data from Fresnel institute, the imaging results are shown in Figure~\ref{fig:diffActFunExp}, where the dataset “dielTM\underline{~}dec8f” are obtained with 36 transmitters and 49 receivers, “FoamTwinDiel” with 18 transmitters and 241 receivers, and both “FoamDielExt” and “FoamDielInt” with 8 transmitters and 241 receivers. The reconstructed permittivity maps demonstrate that, with ReLU, the permittivity tends to be underestimated, and scatterer boundaries are poorly distinguished. With Tanh and Softsign activation functions, strong artifacts appear and the scatterers' permittivity cannot be accurately reconstructed. LeakyReLU achieves reasonable performance under PDNN training, although still inferior to the results with GLOW function in terms of edge preserving and reconstruction accuracy. When learned in FCN scheme, LeakyReLU leads to strong background artifacts. In contrast, the GLOW function produces sharper boundaries, improved contrast, and better fidelity to the ground truth for both PDNN and FCN architectures.
    
    The convergence curves in Figure~\ref{fig:diffActFun},corresponding to the ``FoamDieInt" dataset, shows that the GLOW-based solver converges faster and reaches a lower final error than those using conventional activations. Other activation functions either converge to suboptimal solutions or fail to reach stable convergence, particularly under the FCN scheme. These results confirm the superior stability and accuracy of the proposed activation function.
    
    \begin{figure*}[!ht]
    	\centering
    	\includegraphics[width = \linewidth]{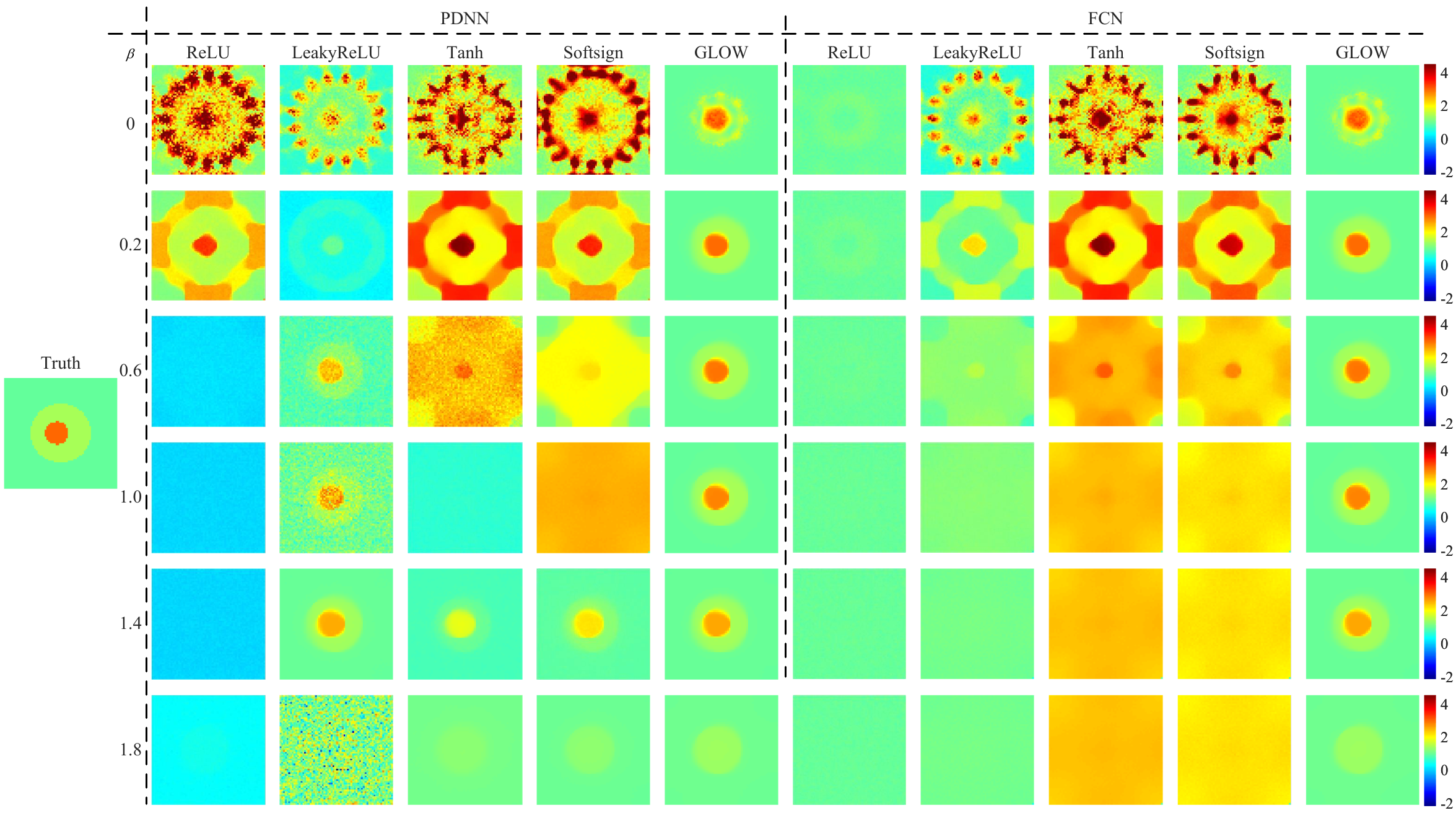}
    	\caption{Imaging examples with the experimental dataset “FoamDielInt” by varying the hyperparameter $\beta$ in loss function while fixing the value of $\alpha$ as 2.}
    	\label{fig:diffActFunExpTV}
    \end{figure*}
    
    Furthermore, the GLOW function exhibits robustness with respect to the hyperparameter $\beta$ in the loss function. Based on the experimental dataset “FoamDielInt”, Figure~\ref{fig:diffActFunExpTV} shows imaging results obtained by varying $\beta$ while fixing $\alpha=2$. Stable and accurate reconstructions are obtained across a wide range of $\beta$ values, except when $\beta=1.8$, where excessive smoothness significantly degrades edge sharpness. Other activations are far more sensitive to $\beta$: for example, LeakyReLU yields good results only when $\beta\in[0.6, 1.4]$, while Tanh and Softsign require $\beta\approx 1.4$ for acceptable reconstructions.
    
    \begin{figure}
    	\centering
    	\includegraphics[width = 0.5\linewidth]{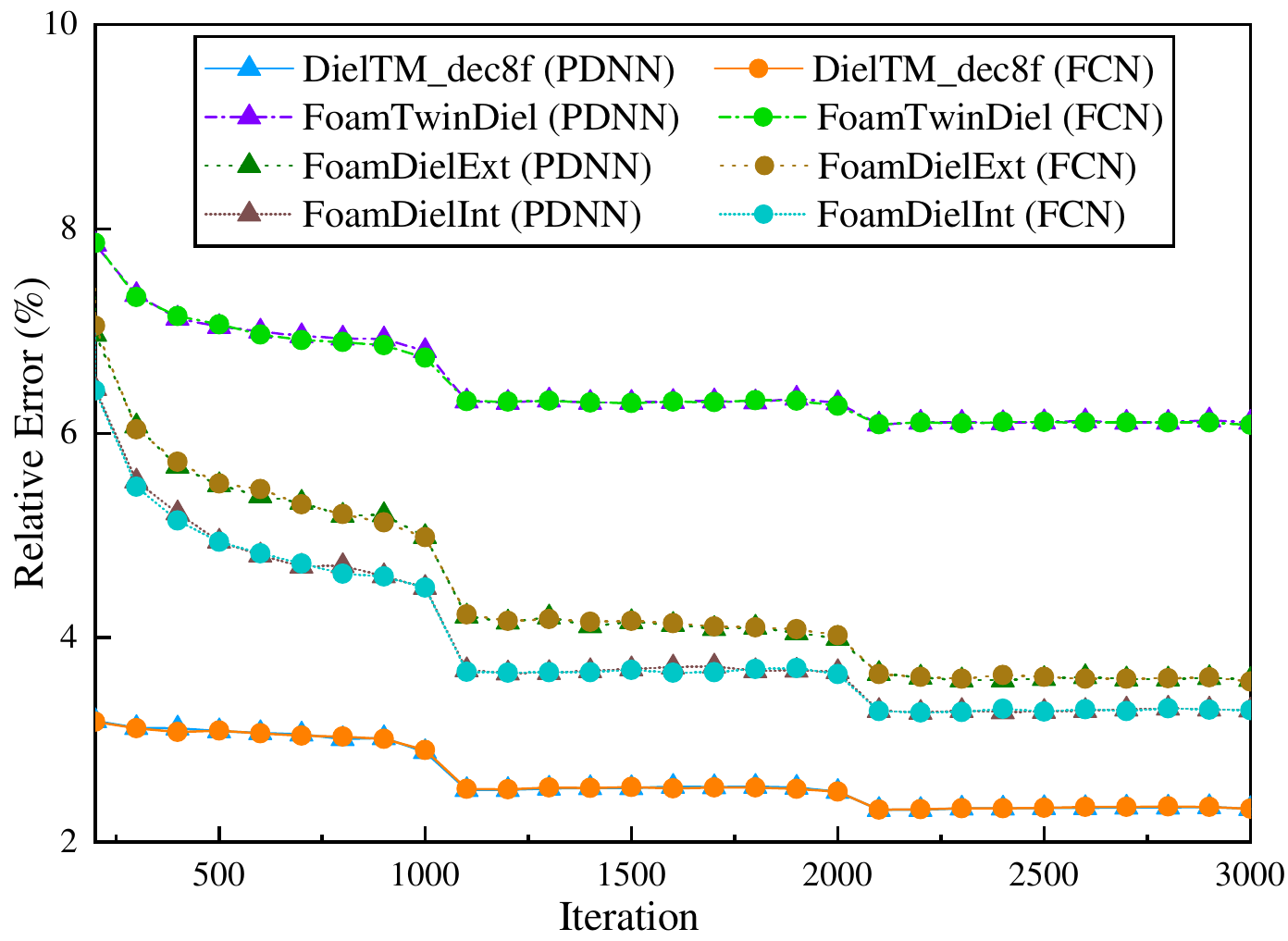}
    	\caption{The convergence curves indicate that with GLOW activation function, the convergence rate is insensitive to the neural network architecture.}
    	\label{fig:diffNet}
    \end{figure}
    \begin{figure*}[!ht]
    	\centering
    	\includegraphics[width = \linewidth]{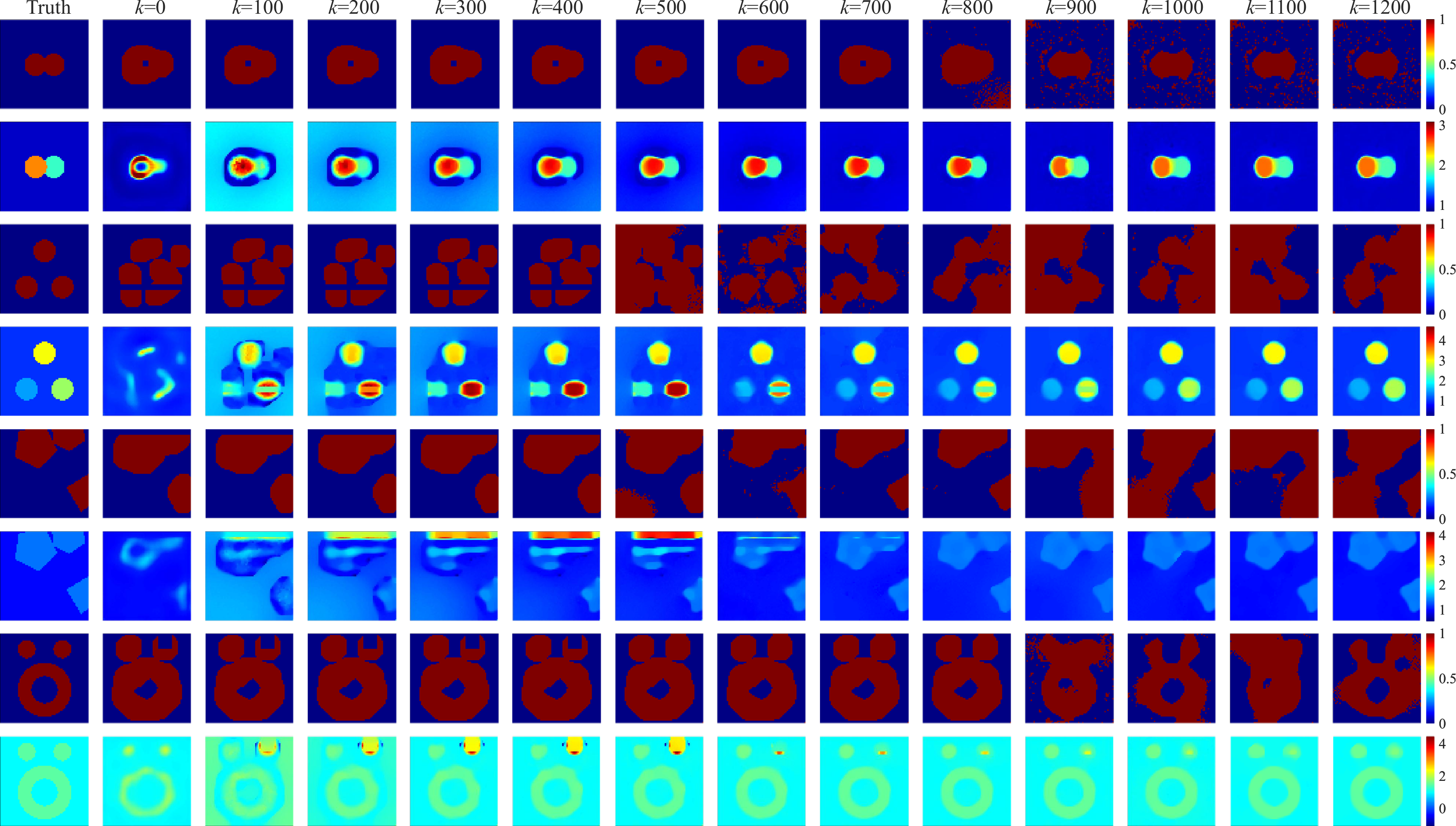}
    	\caption{Dynamic scatter subregion identification and imaging results of the four representative scatterers.}
    	\label{fig:subregionIdentify}
    \end{figure*}
    Importantly, the imaging quality achieved with the GLOW activation is insensitive to the network architecture. As shown in Figure~\ref{fig:diffNet}, both PDNN and FCN solvers converge at nearly identical rates and accuracies. This finding implies that increasing network depth or complexity provides marginal benefit, allowing the IPDNN to employ a single fully connected layer while maintaining high reconstruction fidelity. Consequently, the required memory during training is reduced from 264.49 MB (PDNN) to 16.03 MB, leading to significant computational savings.
    
    \subsection{Dynamic scatter subregion identification}
    \label{subsec:findAreas}
    
    The effectiveness of the proposed dynamic scatter subregion identification method is examined in Figure~\ref{fig:subregionIdentify}, which shows the sequentially updated scatter regions during the iterative inversion process of multiple representative objects. 
    
    For the case with two adjacent circular scatterers, a small region between them is initially missed but is correctly identified after about 800 iterations, resulting in improved permittivity reconstruction accuracy. Similar behavior is observed for the three-circle configuration: missed regions at early iterations are subsequently recovered, while the identified area remains slightly larger than the true scatter due to the conservative update strategy. The imaging results of polygon-shaped and Austria-shaped scatterers further confirm that the dynamic identification mechanism operates reliably even for large and irregular geometries, the main character of which is even observed from the identified subregions.
    
    These observations indicate that the proposed strategy effectively mitigates the risk of missed detections, adaptively refining the region of interest and improving both accuracy and efficiency.
    \begin{figure*}[!ht]
    	\centering
    	\includegraphics[width = \linewidth]{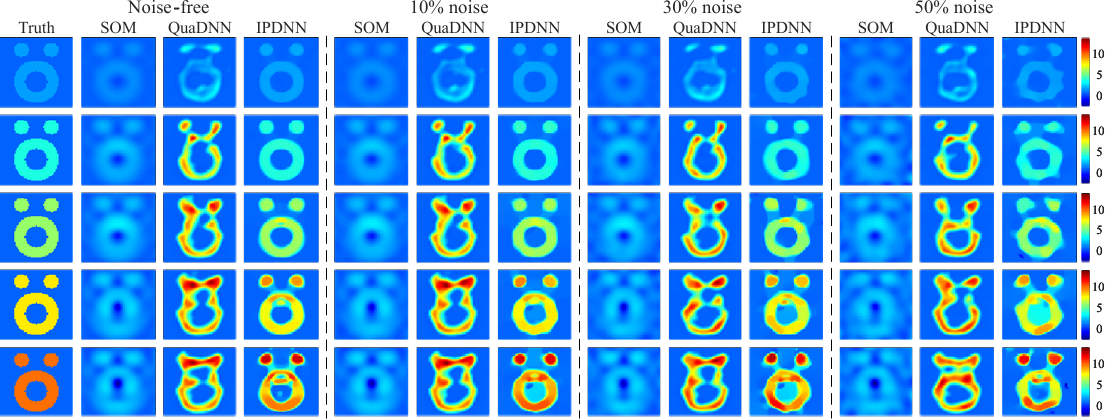}
    	\caption{Test of noise robustness of the proposed IPDNN sovers versus the classical SOM and a data-driven solver QuaDNN \cite{Du2025QuaDNN}.}
    	\label{fig:noisyInfluence}
    \end{figure*}
    \begin{figure*}[!ht]
    	\centering
    	\includegraphics[width = \linewidth]{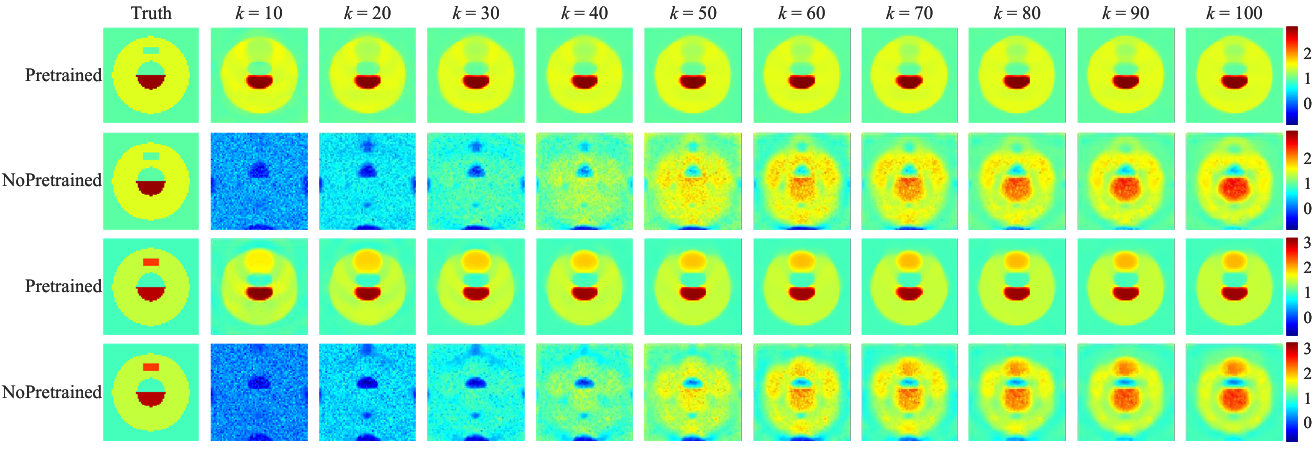}
    	\caption{The intermediate imaging results during iteration with (denoted by ``Pretrained") or without (denoted by ``NoPretrained") the application of transfer learning method show that faster convergence can be achieved by pretraining.}
    	\label{fig:PreTrainedK100}
    \end{figure*}
    \begin{figure}
    	\centering
    	\includegraphics[width = .6\linewidth]{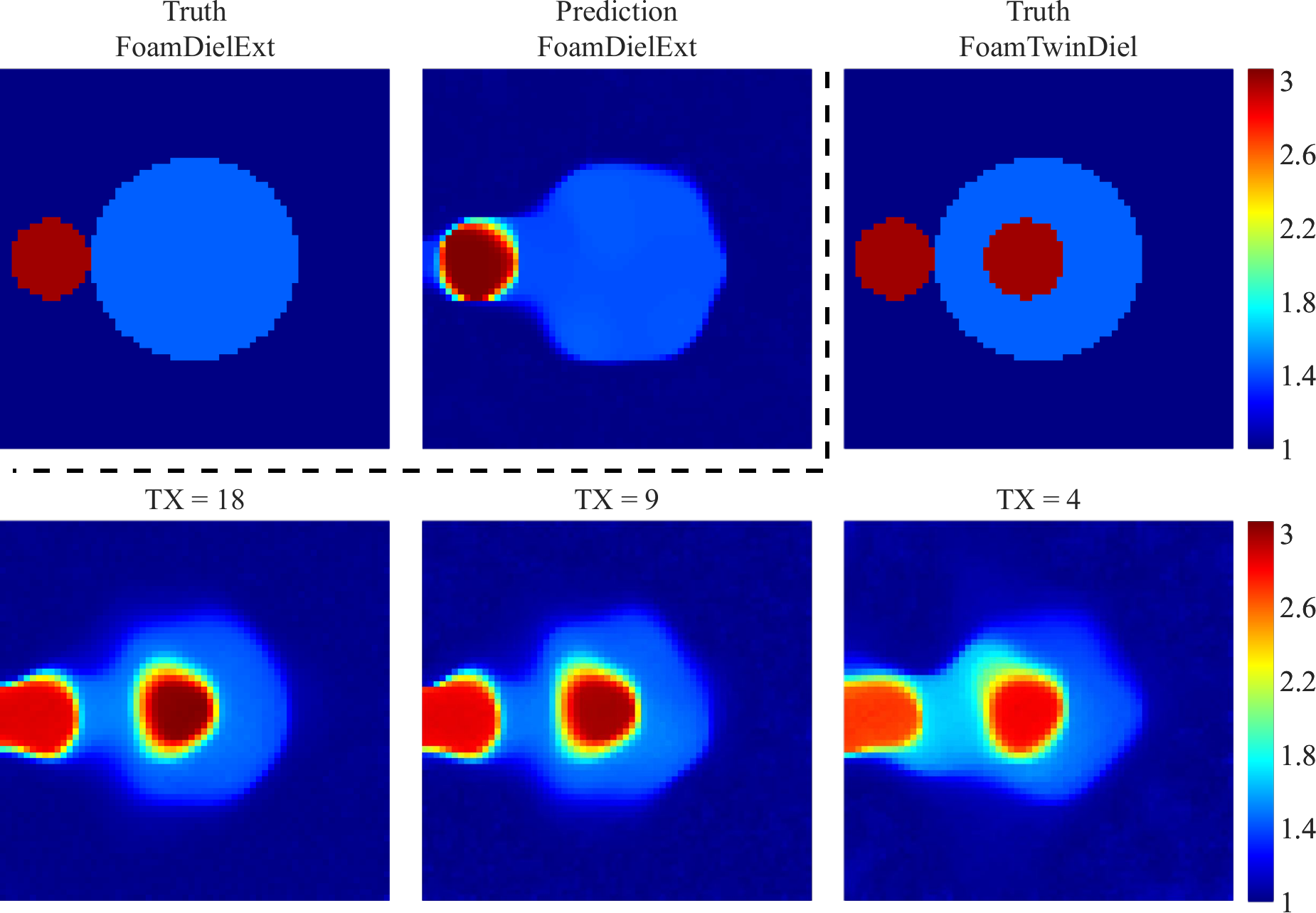}
    	\caption{Predicted relative-permittivity distribution from the solver that pretrained based on the dataset “FoamDielExt”. ``TX" denotes the number of transmitters.}
    	\label{fig:TruthandPretrained}
    \end{figure}

    \subsection{Noise robustness}
    \label{subsec:diffSNR}
    
    To evaluate robustness of the proposed IPDNN solver against measurement noise, additive Gaussian noise is introduced into the scattered fields with noise levels of $10\%$, $30\%$ and $50\%$. The IPDNN is compared with the Subspace Optimization Method (SOM) \cite{chen2010SOM} and a data-driven neural network solver QuaDNN \cite{Du2025QuaDNN}, the latter trained on 2000 digit-like profiles with relative permittivity ranging from 1 to 10. Figure~\ref{fig:noisyInfluence} presents reconstruction results for homogeneous scatterers with true permittivity values of 2, 4, 6, 8, and 10.
    
    The results show that SOM can recover the main character of scatter shape, but tends to underestimate permittivity values and blurs scatter boundaries. QuaDNN is robust to noise but fails to accurately capture the scatter geometric details, especially for high-contrast cases. IPDNN achieves the best overall performance, accurately reconstructing both geometry and permittivity when the noise ratio is below 50\%. Even at 50\% noise, although slight shape distortion appears for high-contrast cases, the main features remain preserved. The above observations demonstrate that the proposed IPDNN exhibits strong noise tolerance and maintains accurate reconstruction for challenging high-contrast conditions.

    \subsection{Transfer learning applications}
    \label{subsec:ApTransLea}
    
    Benefits from transfer learning are validated in a nondestructive testing scenario where a rectangular defect is embedded within a tubular structure. The defect’s relative permittivity is either 1 or 2.5. Since the background material is known, an IPDNN solver is first pretrained using data from the defect-free object, and its weights are subsequently fine-tuned with the defected data.
    
    Figure\ref{fig:PreTrainedK100} compares the imaging results obtained with (``Pretrained") and without (``NoPretrained")
    transfer learning. With pretraining, the defect becomes visible after only 10 iterations, significantly faster than the non-pretrained case. As training proceeds, the defect’s position and size are recovered with higher accuracy. These benefits are consistent across both permittivity values, confirming that pretraining accelerates convergence and enhances reconstruction fidelity.
    
    The same concept is applied to experimental data. As shown in Figure~\ref{fig:TruthandPretrained}, the ``FoamDielExt"
    dataset is used to pretrain the solver for ``FoamTwinDiel” imaging. The pretrained model achieves accurate reconstruction using all 18 transmitters and remains effective even when the number of transmitters is reduced to 9 or 4. Although the low-contrast scatterer becomes slightly distorted when using only 4 transmitters, the shapes and permittivity of the stronger scatterers are still correctly retrieved. This demonstrates that transfer learning can reduce data requirements without compromising performance much.
    
    \section{Conclusions}
    \label{sec:conclusions}
    This paper presents an improved physics-driven neural network (IPDNN) for solving electromagnetic inverse scattering problems. The proposed GLOW activation function enhances nonlinear feature extraction and ensures stable convergence with a compact, single-layer network. A dynamic scatter subregion identification algorithm further improves reconstruction accuracy and efficiency by adaptively refining the computational domain. Additionally, transfer learning enables fast adaptation to practical applications such as nondestructive testing, combining the physical interpretability of iterative algorithms with the rapid inference of data-driven models.
    
    Extensive numerical and experimental results verify that the IPDNN achieves superior imaging accuracy, noise robustness, and computational efficiency compared with existing methods. Although small-scale defect imaging remains challenging, future work will focus on integrating high-resolution feature extraction modules and multi-frequency training strategies to enhance defect localization performance.

	\bibliographystyle{unsrt}
	\bibliography{DNN}
\end{document}